\documentclass[10pt, a4paper]{article}
\usepackage{lrec2022} 
\usepackage{multibib}
\newcites{languageresource}{Language Resources}

\usepackage{graphicx}
\usepackage{tabularx}
\usepackage{soul}
\usepackage{here}
\usepackage{color}
\usepackage{soul}
\usepackage{booktabs}
\usepackage{tabularx}
\usepackage{graphicx}
\usepackage{multirow}
\usepackage{etaremune}
\usepackage{comment}
\graphicspath{ {./images/} }
\usepackage{titlesec}
\titleformat{\section}{\normalfont\large\bfseries\center}{\thesection.}{1em}{}
\titleformat{\subsection}{\normalfont\SmallTitleFont\bfseries\raggedright}{\thesubsection.}{1em}{}
\titleformat{\subsubsection}{\normalfont\normalsize\bfseries\raggedright}{\thesubsubsection.}{1em}{}
\renewcommand\thesection{\arabic{section}}
\renewcommand\thesubsection{\thesection.\arabic{subsection}}
\renewcommand\thesubsubsection{\thesubsection.\arabic{subsubsection}}

\usepackage{epstopdf}
\usepackage[utf8]{inputenc}

\usepackage{hyperref}
\usepackage{xstring}

\usepackage{color}

\title{Proficiency Matters Quality Estimation in Grammatical Error Correction}

\name{Yujin Takahashi$^1$, Masahiro Kaneko$^2$, Masato Mita$^{3,1}$, Mamoru Komachi$^1$} 

\address{$^1$Tokyo Metropolitan University, 
        $^2$Tokyo Institute of Technology, 
        $^3$RIKEN \\
         takahashi-yujin@ed.tmu.ac.jp,
         masahiro.kaneko@nlp.c.titech.ac.jp,
         masato.mita@riken.jp,
         komachi@tmu.ac.jp\\}

\abstract{
This study investigates how supervised quality estimation (QE) models of grammatical error correction (GEC) are affected by the learners' proficiency with the data.
QE models for GEC evaluations in prior work have obtained a high correlation with manual evaluations.
However, when functioning in a real-world context, the data used for the reported results have limitations because prior works were biased toward data by learners with relatively high proficiency levels.
To address this issue, we created a QE dataset that includes multiple proficiency levels and explored the necessity of performing proficiency-wise evaluation for QE of GEC.
Our experiments demonstrated that differences in evaluation dataset proficiency affect the performance of QE models, and proficiency-wise evaluation helps create more robust models.
 \\ \newline  }

\begin{document}

\maketitleabstract


\section{Introduction}
Grammatical error correction (GEC) refers to the task of correcting a variety of grammatical errors in text written by non-native speakers learning a language.
Thus, the main application of a GEC system is to assist language learners with their writing.
The performance of GEC systems has been improving, but there is still room for improvements.
The precision of the state-of-the-art system is approximately $75$\% and recall is below $50$\% on the CoNLL-2014 test set~\cite{Stahlberg+2021}.
Therefore, it is difficult for learners to judge whether they can trust the GEC system's output, as it may be potentially misleading in some cases.

The quality estimation (QE) of GEC, which is the task of estimating the quality of GEC system outputs, has attracted attention as a key method to address the aforementioned difficulties in real-world scenarios~\cite{heilman+2014,chollampatt+2018,Asano+2017,Yoshimura+2020}.
Having quality estimates of the system's output can help learners and instructors decide whether to adopt or ignore the system's correction based on the estimation of its quality.
Recently, it has been reported that QE using BERT~\cite{Devlin+2019} was highly correlated with human evaluations~\cite{Yoshimura+2020}.

However, the current evaluation in QE of GEC is unfortunately insufficient because it is restricted to a corpus limited to certain proficiency levels.
For example, the data used in prior works~\cite{Asano+2017,Yoshimura+2020} were biased toward data generated for learners with relatively high proficiency levels such as CoNLL-2014~\citelanguageresource{Ng+2014}.
By contrast, it is assumed that the proficiency level of language learners varies from beginners to advanced learners in real-world use cases.
Given the nature of the QE task, QE systems need to be able to estimate quality with high accuracy for sentences written by learners with low proficiency, who may have difficulty in trusting the system’s output. 
However, the current evaluation cannot address this issue.
If research is conducted based on such a limited evaluation, the QE systems will be overfitted to a specific type or genre of written English.

To overcome these limitations and improve the real-world applicability of GEC with QE, we present a new dataset, \textbf{ProQE:} \textbf{Pro}ficiency-wise \textbf{Q}uality \textbf{E}stimation dataset, that includes multiple proficiency levels designed for proficiency-wise evaluation.
Using our dataset, we first explore the necessity of proficiency-wise evaluation based on the hypothesis that it is insufficient to evaluate only a specific proficiency level as a methodology for evaluating QE models, and then reveal how QE models are affected by the learner's proficiency.

Thus, the contributions of this work are three-fold:
(1) We propose a novel dataset designed for proficiency-wise evaluation in QE of GEC to accommodate the detailed analysis of the impact of learner's proficiency on models' performance;
(2) Using this dataset, we demonstrated that results varied when the evaluation data contained different proficiency levels, and that we could create more robust QE models based on our findings;
(3) Our dataset will be made public to allow researchers in the community to easily conduct proficiency-wise evaluations.


\section{Related Work}

We were motivated by the issue of robustness in the GEC community.
Prior works focused on improving the performance of GEC systems on CoNLL-2014~\cite{Ng+2014}.
However, it has been recently reported that a single corpus evaluation is insufficient because the input of GEC (i.e., the learner text) is diverse with respect to the learner's proficiency and native language (L1), and the system performance is sensitive to these factors~\cite{Mita+2019}.
Additionally, and with similar motivation, the BEA-2019 Shared task~\citelanguageresource{Bryant+2019} was held and provided the research community with proficiency-wise data for more robust evaluations.
We conjecture that a similar situation occurs in QE of GEC datasets.

The evaluation methods for GEC include reference-based and reference-less evaluations.
Reference-based metrics, such as Max Match~\cite{Dahlmeier+2012}, ERRANT~\cite{Bryant+2017}, and GLEU~\cite{Napoles+2015}, face a fundamental problem in that it is difficult to include all possible references, even if they are grammatically correct~\cite{Choshen+2018}.
As a result, prior works have proposed reference-less evaluations to address these limitations.
\newcite{Napoles+2015} first presented a reference-less evaluation using grammatical error detection tools and linguistic features. 
\newcite{Asano+2017}, alternatively, combined three sub-metrics (grammaticality, fluency, and meaning preservation) and achieved a higher correlation with manual evaluations than did the previous reference-based metrics.
Moreover, \newcite{Yoshimura+2020} optimized each sub-metric for manual evaluation and obtained larger improvements using BERT-based QE models.
In GEC, reference-less evaluations are a part of QE because it estimates the quality of the system’s output without requiring gold-standard references. 


\section{The ProQE dataset}
\label{sec:dataset}

\paragraph{Dataset Design}
To analyze the impact of the learners' proficiency level on the QE model, we require a dataset that is differentiated by proficiency level. 
To serve this purpose, we considered that it was desirable that the data with proficiency information be sourced from the CEFR\footnote{https://www.cambridgeenglish.org/exams-and-tests/cefr/}, an international index for assessing language proficiency.
Hence, we selected the Write \& Improve (W\&I) and LOCNESS~\citelanguageresource{Bryant+2019,Granger+1998} datasets because they meet our requirements, and their use is prevalent in the GEC community.
W\&I includes non-native English students' writing across three different proficiency levels: beginner (A), intermediate (B), and advanced (C). 
Furthermore, LOCNESS contains essays written by native (N) English students. 
As requirements of system outputs, we considered that the systems must be diverse and commonly
used since we need to ensure the applicability in a practical use case.
Hence, we adopted five diverse and commonly used GEC systems, (SMT~\cite{SMT:18}, RNN~\cite{RNN:15}, CNN~\cite{CNN:18}, Transformer~\cite{SAN:17}, and Transformer with a copy mechanism~\cite{CPY:19}), to obtain system outputs.

\begin{figure}[t]
    \centering    
    \includegraphics[width=0.43\textwidth]{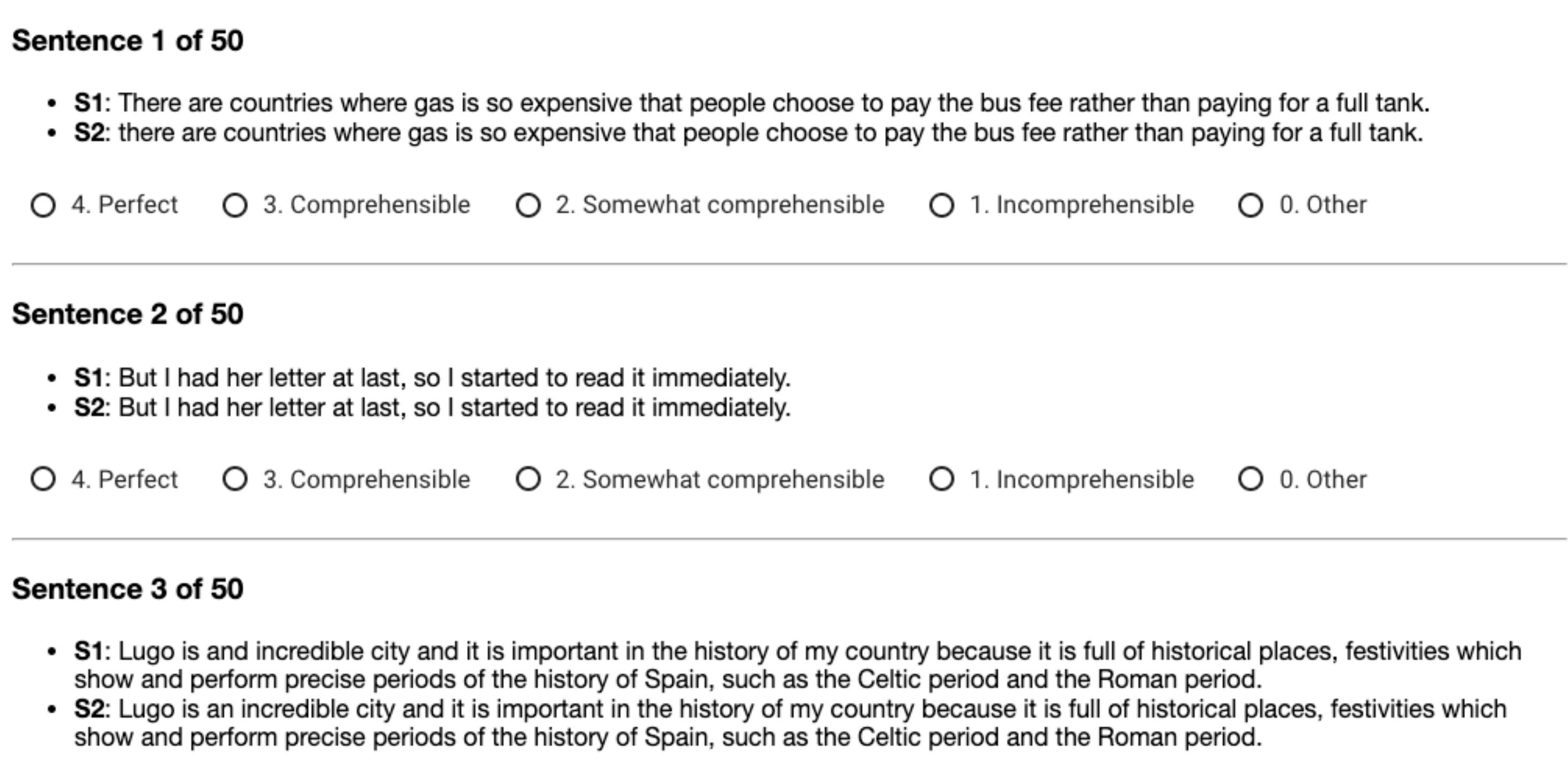}
    \caption{Screenshot of actual task on Amazon Mechanical Turk.}
    \label{fig:task_screen}

\vspace{0.25cm}
\footnotesize
\begin{tabular}{p{\columnwidth}}
Please rate each of the following 50 sentences.
You should evaluate using two-sentences; an original sentence (S1) and a corrected sentence (S2).
Please rate the S2 (a corrected sentence) on a scale of 5 overall.
The evaluation criteria for each label are as follows. \\
\end{tabular}
\begin{tabular}{cp{6.7cm}}
\toprule
label & criteria \\
\midrule
4 & S2 corrected all errors completely. (S2 has perfect grammar.) \\
3 & S2 corrected serious errors and fixed many minor errors. (S2 has one or more minor grammatical errors.) \\
2 & S2 corrected serious errors but contains minor errors and/or incorrect minor corrections. \\
1 & S2 did not correct serious and minor errors and/or contains serious incorrect corrections. \\
0 & S1 is an incomplete sentence. (S1 cannot be corrected.) \\
\bottomrule
\end{tabular}
\scriptsize
Note: S1 has a possibility of containing no grammatical errors.
\caption{Task description.}
\label{fig:task_description}
\end{figure}

\begin{figure}[t]
    \centering    
    \includegraphics[width=0.43\textwidth]{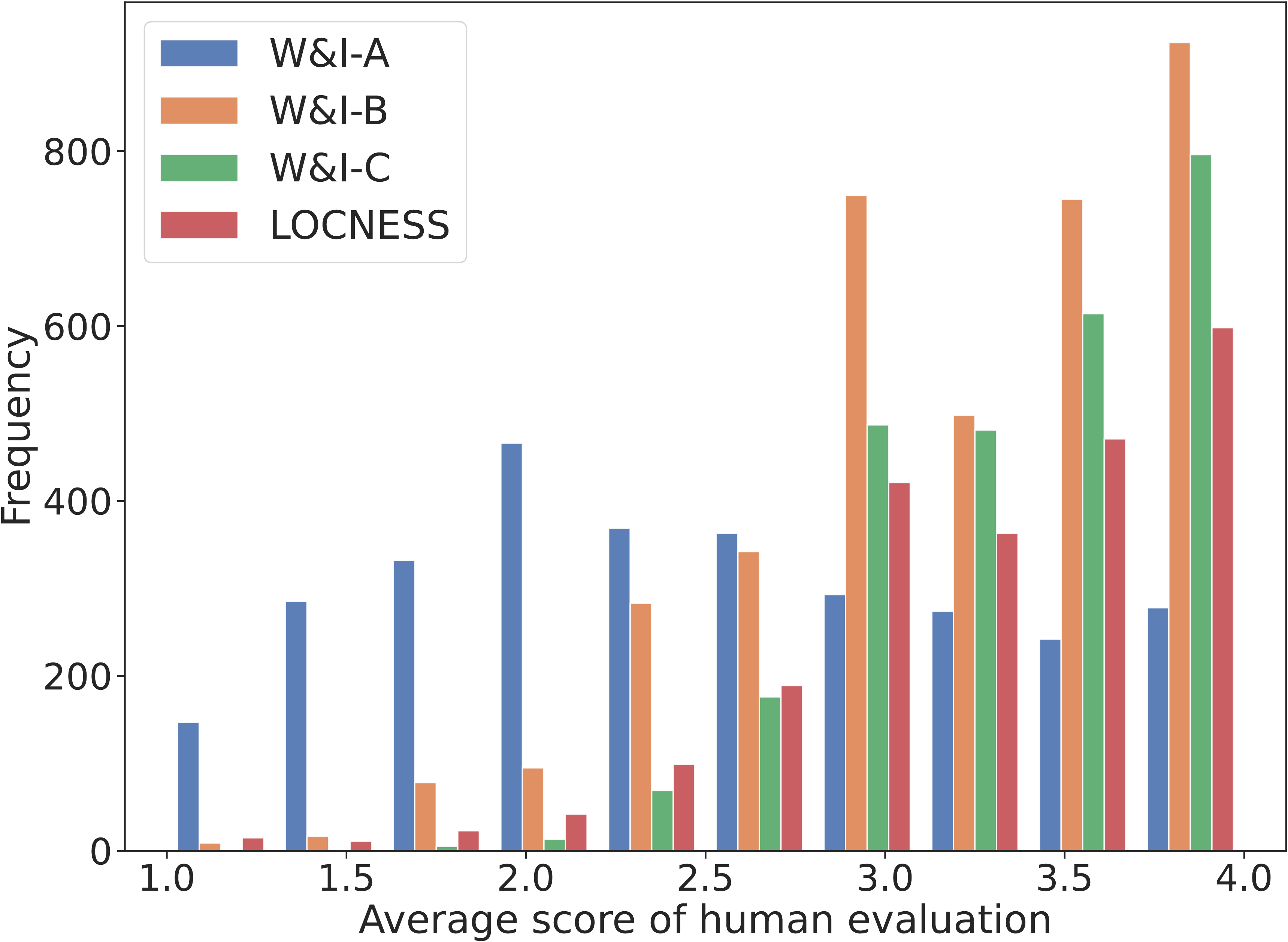}
    \caption{Histogram of each average score.}
    \label{fig:dist_dataset}
\end{figure}

\begin{table}[t]
    \centering
    \small
    \begin{tabular}{rl}
        \toprule
            & example sentence \\
        \midrule \midrule
            source &  I live \textcolor{red}{at} Patras a big city in Greece. \\
            output &  I live \textbf{in} Patras big \textcolor{red}{cities} in Greece. \\
            scores &  2, 1, 1 (avg. score: 1.33) \\
        \midrule
            source & Let me allow to enter \textcolor{red}{in to} your cabin. \\
            output & Let me allow \textbf{you} to enter your cabin. \\
            scores  & 4, 4, 4 (avg. score: 4.00) \\
        \bottomrule
    \end{tabular}
    \caption{Example annotations}
    \label{tab:example_annot}
\end{table}

\begin{table}[t]
    \centering
    \small
    \begin{tabular}{lccccc}
    \toprule
        level
        & A & B & C & N & avg. \\
    \midrule \midrule
        sent. & 3,049 & 3,740 & 2,644 & 2,233 & 2,917\\
    \bottomrule
    \end{tabular}
    \caption{Number of sentences in each created dataset.}
    \label{tab:detail_dataset}
\end{table}

\begin{table*}[t]
    \centering
    \small
    \begin{tabular}{l rrrr rrrr}
        \toprule
        & \multicolumn{4}{c}{Pearson's correlation} 
        & \multicolumn{4}{c}{Spearman's rank correlation} \\
        \cmidrule(lr){2-5} \cmidrule(lr){6-9}
        \multicolumn{1}{c}{Model}
        & \multicolumn{1}{c}{A} & \multicolumn{1}{c}{B}
        & \multicolumn{1}{c}{C} & \multicolumn{1}{c}{N}
        & \multicolumn{1}{c}{A} & \multicolumn{1}{c}{B}
        & \multicolumn{1}{c}{C} & \multicolumn{1}{c}{N} \\ 
        \midrule 
        \textsc{Beginner}        
        & \textbf{0.70\textpm.05} & 0.54\textpm.03          & 0.51\textpm.05          & 0.60\textpm.05 
        & \textbf{0.69\textpm.05} & 0.55\textpm.03          & 0.52\textpm.05          & 0.60\textpm.05 \\
        \textsc{Intermediate}
        & 0.63\textpm.03          & 0.58\textpm.06          & 0.52\textpm.06          & 0.57\textpm.05
        & 0.65\textpm.04          & 0.58\textpm.05          & 0.52\textpm.06          & 0.57\textpm.04 \\
        \textsc{Advanced}
        & 0.63\textpm.02          & 0.52\textpm.04          & 0.52\textpm.05          & 0.58\textpm.05 
        & 0.66\textpm.05          & 0.53\textpm.02          & 0.52\textpm.05          & 0.58\textpm.02 \\
        \textsc{Native}
        & 0.59\textpm.05          & 0.54\textpm.05          & 0.49\textpm.05          & \textbf{0.61\textpm.04}
        & 0.63\textpm.07          & 0.56\textpm.03          & 0.50\textpm.07          & 0.60\textpm.03 \\
        \midrule
        \textsc{RandSamp}
        & 0.59\textpm.02          & 0.53\textpm.06          & 0.51\textpm.04          & 0.57\textpm.05
        & 0.63\textpm.04          & 0.54\textpm.05          & 0.51\textpm.04          & 0.57\textpm.05 \\
        \textsc{Mixed}
        & 0.68\textpm.06          & \textbf{0.58\textpm.05} & \textbf{0.57\textpm.03} & 0.60\textpm.02
        & 0.69\textpm.06          & \textbf{0.60\textpm.05} & \textbf{0.57\textpm.05} & \textbf{0.61\textpm.05} \\
        \midrule \midrule
        \textsc{Mixed+Tag }
        & \textbf{0.72\textpm.06} & \textbf{0.61\textpm.04} & \textbf{0.60\textpm.05} & \textbf{0.62\textpm.02}
        & \textbf{0.71\textpm.06} & \textbf{0.63\textpm.02} & \textbf{0.60\textpm.04} & \textbf{0.63\textpm.03} \\
        \bottomrule
    \end{tabular}
   \caption{Result of the proficiency-wise evaluation.}
    \label{tab:ft_result}
\end{table*}

\paragraph{Annotation}
For annotation, we used Amazon Mechanical Turk\footnote{https://www.mturk.com/} and recruited three native English speakers as annotators.
First, we assigned unique sentence pairs except for duplicate corrected sentences to the annotators.
Then, the annotators used both the source and the corrected sentence to evaluate the quality of the correction.
Figures~\ref{fig:task_screen} and \ref{fig:task_description} show an actual working example and the task description.
Note that we did not annotate each of the sub-metrics unlike~\newcite{Yoshimura+2020} because their report showed substantial weighting of one metric and its high correlation with the holistic score.
Therefore, we created five-scale evaluation criteria (4: Perfect, 3: Comprehensible, 2: Somewhat comprehensible, 1: Incomprehensible, 0: Other) to evaluate the overall correction quality. 
Figure~\ref{fig:dist_dataset} shows the distribution of the evaluation scores.
The beginner level data were not biased for every score, but for the intermediate and higher levels, the higher scores were prevaent.
For reference, we provide examples of annotations in Table~\ref{tab:example_annot}.
Using the above procedure, we obtained
3,049 sentences for A, 3,740 sentences for B, 2,644 sentences for C, and 2,233 sentences for N.
Table~\ref{tab:detail_dataset} lists the number of sentences for each created dataset.

\paragraph{Ethical Consideration}
\label{adx:ethical}

The annotator was given two sentences in a question, one before and one after the correction, and evaluated the quality of the correction.
Each task included 50 questions; we paid \$$1.3$ per task, which resulted in an average hourly wage of \$$7.8$. 
The entire corpus collection took approximately five weeks.


\section{Proficiency-wise Evaluation}
The main objective of our experiment here is to investigate the necessity of proficiency-wise evaluation based on the hypothesis that it is insufficient to evaluate only a specific proficiency level as a methodology for evaluating QE models.
Thus, following the methodology introduced by \newcite{Mita+2019}, we prepare a wide variety of QE models and examine whether the performance ranking of the QE models changes between each proficiency level.

\subsection{Configurations}

\paragraph{Evaluation}
We report Pearson's correlation coefficients and Spearman's rank correlation coefficients.
All scores were evaluated using five-fold cross-validation because the evaluation results were highly dependent on the seed value of the data division.

\paragraph{Baseline Models}
We adopted BERT used in \newcite{Yoshimura+2020}, which achieved the highest correlation with the manual evaluations, as the base architecture for our baseline models.
In these experiment, we arranged six models in terms of two aspects: (1) writer's proficiency and (2) data size.
Specifically, for the first, we employed a total of four different models by fine-tuning BERT for each proficiency level; \textsc{Beginner}, \textsc{Intermediate}, \textsc{Advanced}, and \textsc{Native}.
To avoid differences in the number of sentences, we unified the number of sentences to 2,233 based on the dataset with the fewest number of sentences\footnote{For datasets with more than that many sentences, we randomly selected 2,233 sentences.}.
For the second, we employed two models: one fine-tuned with the combined data of all the proficiency levels (\textsc{Mixed}), and the other fine-tuned with the data randomly sampled from the combined data so that the data size is the same as other models except \textsc{Mixed} (\textsc{RandSamp}).
We performed a grid search for hyperparameters to maximize Pearson's correlation coefficient.

\subsection{Result}
Table~\ref{tab:ft_result} shows the result of the proficiency-wise evaluation.
The evaluation results reveal that the models' rankings considerably vary depending on the proficiency, indicating that it is insufficient to evaluate only a specific proficiency level as a methodology for evaluating QE models.

\paragraph{Difference in Proficiency}
The top group in Table~\ref{tab:ft_result} shows the effect of differences in proficiency levels.
Experimental results show that each QE model performed better correlations when the learners' proficiency levels in the data were consistent at each stage of fine-tuning and evaluation. 
Except for the evaluation data at the beginner level, there were no significant differences in the correlations.
By contrast, we confirmed that there was a large difference between \textsc{Beginner} and \textsc{Native} in the beginner-level evaluation data (e.g., 0.70\textpm.05 vs. 0.59\textpm.05 in Pearson's correlation).
We provide the detailed analysis on this observation in Section~\ref{subsec:dist_pred}.

\paragraph{Importance of Data Size}
The second group in Table~\ref{tab:ft_result} shows the effect of differences in data size.
Although the performance of \textsc{RandSamp} was the lowest, or nearly the lowest score for all of the evaluation data,
the performance of \textsc{Mixed}  was competitive for almost all of the evaluation data. 
\textsc{Mixed} obtained improvement by increasing the data size, but it did not outperform the best model for each proficiency level in some cases.


\begin{figure}[t]
    \centering    
    \includegraphics[width=0.40\textwidth]{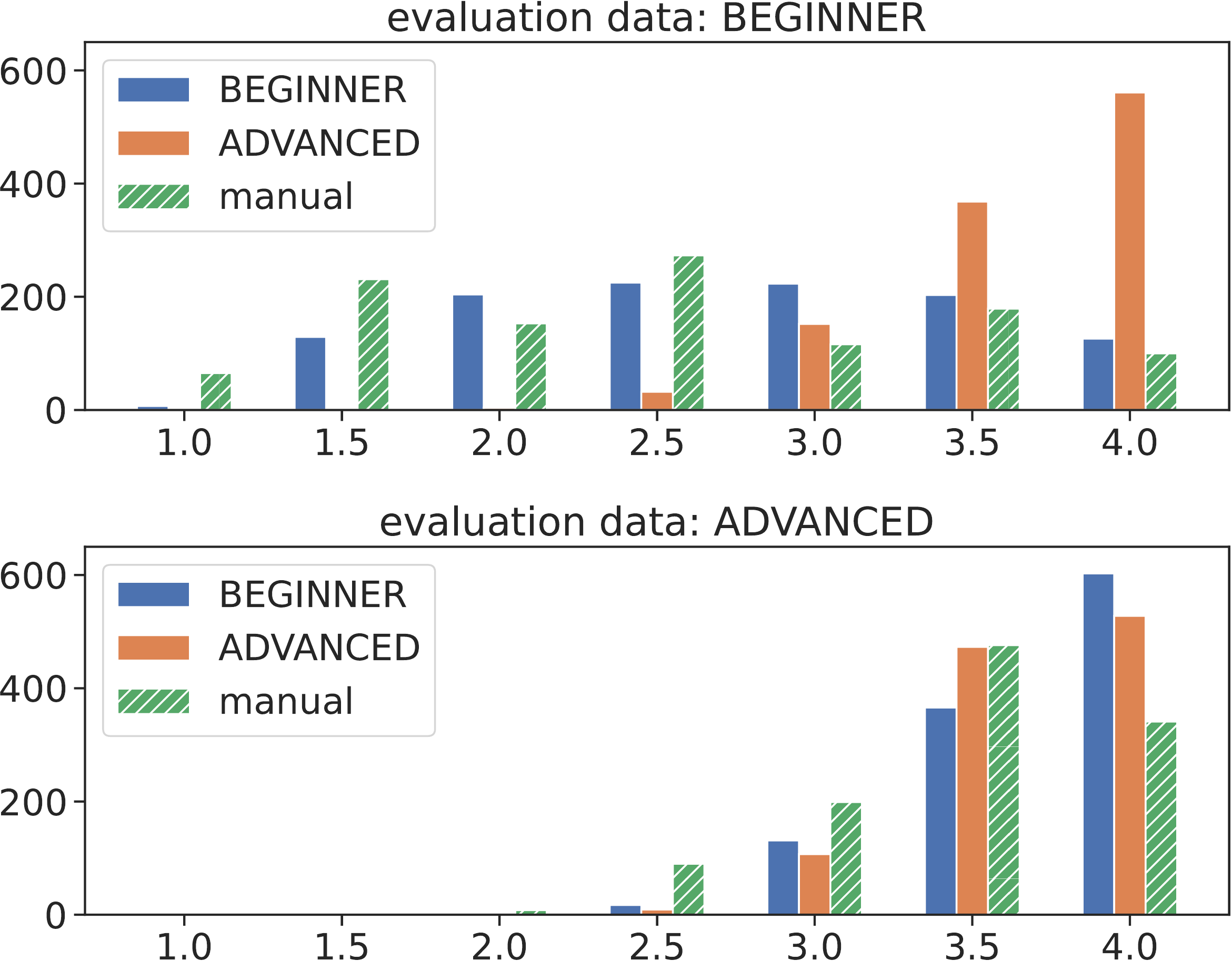}
    \caption{Distribution of prediction scores by QE.}
    \label{fig:dist_predction}
\end{figure}

\section{Discussion and Analysis}

\subsection{On the Impact at the Beginner Level}
\label{subsec:dist_pred}
We found that the impact of the proficiency was significant especially in the beginner level. 
One reason for this effect might be related to the distribution of the scores of the manual evaluation because the distribution differed significantly between the beginner and other proficiency levels (Figure~\ref{fig:dist_dataset}).
Therefore, we assumed that the QE models fine-tuned on data from the high proficiency level tended to produce high prediction values, even for low-quality inputs.

To verify this hypothesis, we confirmed the distribution of the manual evaluation values and the predicted values of the QE model for the evaluation data of the two proficiency levels: beginner and advanced (Figure~\ref{fig:dist_predction}).
In the case of the beginner level, \textsc{Beginner} had a distribution similar to that of the manual evaluation. 
Still, \textsc{Advanced} only produced predictions of 2.5 or higher.
By contrast, for the evaluation data from the advanced level, the QE model fine-tuned at any proficiency levels showed almost no significant difference.
Thus, the bias in the distribution of the manual evaluation caused the QE model to produce overestimations for low-quality input.

\begin{figure}[t]
    \centering    
    \includegraphics[width=0.45\textwidth]{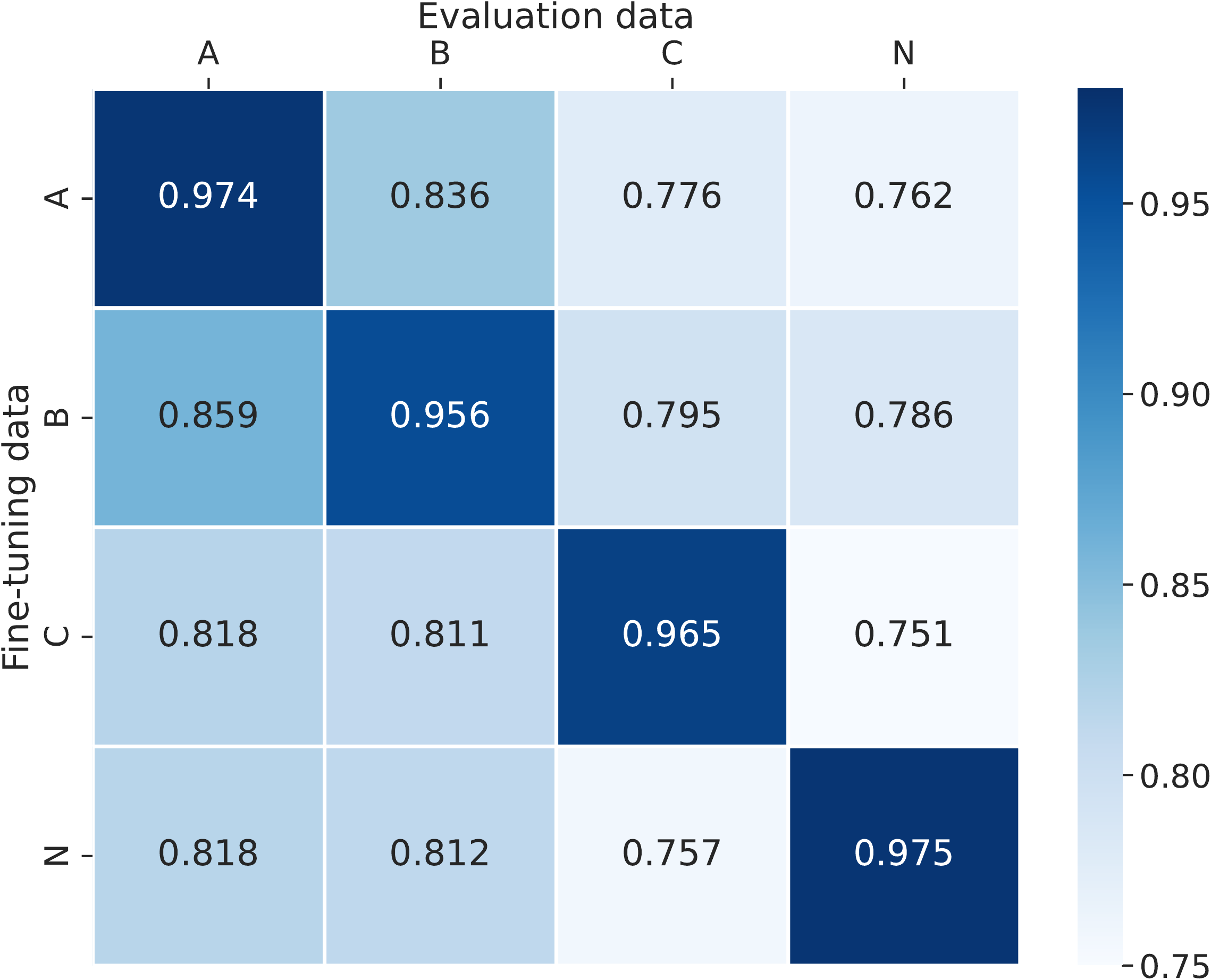}
    \caption{Overlap of vocabulary between the fine-tuning and evaluation data.}
    \label{fig:overlap_vocab}
\end{figure}

\subsection{Proficiency and Vocabulary Overlap}
Although the distributions of manual evaluation from intermediate to native were similar, 
the correlation tended to be slightly higher when the proficiency level matched the fine-tuning and evaluation stages.
We presumed that this is because of the influence of vocabulary differences caused by increased proficiency levels. 
Hence, we investigated the overlap of words of each data.

Figure \ref{fig:overlap_vocab} shows the overlap of the vocabulary between fine-tuning and evaluation data. 
Because each model increased the vocabulary size of evaluation data during the fine-tuning phase, 
we assumed that this contributed to slight differences. 
Thus, one reason for this slight difference in the correlation results may be differences in vocabulary.

\subsection{Seeking a More Robust QE Model}

From our analysis, we found that proficiency information and data size contributed to improving the performance.
Based on our findings, we examined whether adding proficiency information to the \textsc{Mixed} model could improve its performance.
Specifically, we created a setting, \textsc{Mixed+Tag}, in which each sentence was prefixed with a proficiency tag (e.g., \verb|[A]|) for fine-tuning.

The bottom of Table~\ref{tab:ft_result} shows the results of this settings.
We confirmed improved correlations with the evaluation data for all proficiency.
In particular, we verified an increase in beginner level evaluation data, which slightly improved for the others.
This result demonstrated that proficiency-wise evaluation could help create robust QE models.


\section{Conclusions}
In this study, we performed proficiency-wise evaluation using the ProQE dataset and presented the necessity of proficiency-wise evaluation for QE of GEC. 
Furthermore, based on the results, we showed it to help create robust QE models.
To facilitate more research on GEC with QE, we will make the dataset and software freely available.


\section{Bibliographical References}\label{reference}

\bibliographystyle{lrec2022-bib}
\bibliography{lrec2022}

\section{Language Resource References}
\label{lr:ref}
\bibliographystylelanguageresource{lrec2022-bib}
\bibliographylanguageresource{languageresource}

\end{document}